# Exploring semantically-related concepts from Wikipedia: the case of SeRE[1]

Daniel Hienert

*GESIS – Leibniz-Institute for the Social Sciences, Cologne, Germany*

Dennis Wegener

*GESIS – Leibniz-Institute for the Social Sciences, Cologne, Germany*

Siegfried Schomisch

*GESIS – Leibniz-Institute for the Social Sciences, Cologne, Germany*

**Abstract:** In this paper we present our web application SeRE designed to explore semantically related concepts. Wikipedia and DBpedia are rich data sources to extract related entities for a given topic, like in- and out-links, broader and narrower terms, categorisation information etc. We use the Wikipedia full text body to compute the semantic relatedness for extracted terms, which results in a list of entities that are most relevant for a topic. For any given query, the user interface of SeRE visualizes these related concepts, ordered by semantic relatedness; with snippets from Wikipedia articles that explain the connection between those two entities. In a user study we examine how SeRE can be used to find important entities and their relationships for a given topic and to answer the question of how the classification system can be used for filtering.

## 1. Introduction

Wikipedia collects encyclopedic knowledge in a very compact way with a quality controlled by an open community of wikipedia contributors. Entities such as individuals, institutions or abstract concepts are included as separate articles into Wikipedia following a relevance check. The content of the articles is constantly monitored by the community. As a result, important entities and facts about them are represented in Wkipedia fairly accurately and with a fair level of detail represented in a structured and compact fashion. Due to a categorisation system, imposed in the process of article creation, the entities end up by being placed into an overall hierarchy. This is in contrast to the general web, where a large amount of information is available but not explicitly organized and quality checked. Developers in, for example, a knowledge base field take advantage of the rich content of Wikipedia by extracting factual and conceptual knowledge and making it available in a machine 'understandable' form. This creates the basis for the next step at which it is possible to infer knowledge, but also to make this knowledge available to the user in an interactive fashion.

---

1  *SeRE* is a research prototype to explore relationships beween concepts in supporting resource discovery in the Web environment. The Wikipedia and DBpedia are used as databases and are queried live. The prototype was built by D. Hienert as a research project, and it is described for the first time in this article.

In our application SeRE[2] (Semantic Relatedness Explorer) we take advantage of (1) the explicit choice of entities, (2) quality checked facts in full text, and (3) the categorisation system for the selection and computation of important related entities for an arbitrary search term. Unlike semantic search engines, we use the full text of Wikipedia articles for the calculation of semantic relatedness and for the display of relationships. This makes it possible to explore important concepts for a known or unknown subject and the display of textual connections between these concepts.

The structure of this paper is as follows: In Section 2 we present related work in the knowledge base field, the role of visualization in classification and current visualization systems; Section 3 contains presentation of our own approach to computing semantically related concepts from Wikipedia; Section 4 provides an overview of the user interface and in Section 5 we present a user study in which we analyse the basic goals of classification visualization with our own application. Finally, Section 6 contains summary and concluding remarks.

**2. Related Work**

Eppler & Stoyko (2009) describe the role of classification in research and how graphical representations can support the exploration process. Graphics can be used to show the grouping hierarchy, visualize the attributes of items and show relations among groups in an easily understandable and interactive fashion. For the visualization part the authors distinguish between four types of visual classification: layers, trees, compilations and configurations, which emphasize the hierarchical or relational aspect differently. Merčun & Žumer (2010) give an overview of online visual tools for exploration and discovery. These tools are based on the principles of (1) discovery and serendipity (compare for example Foster & Ford, 2003) and (2) on the exploratory search process (Marchionini, 2006). The first principle describes the unexpected discovery of information while searching, the second describes a search process emphasizing learning and investigation steps. Visualizations of a classification structure can be used, especially in the early stages of the research process, to get an overview of the area and to make comparisons of groups and concepts inside the topic (Eppler & Stoyko, 2009). A current system used for the visualization of categorised, related concepts from Wikipedia is *EyePlorer*[3] (Ritschel, Pfeiffer & Mende, 2010). Here, the user can enter a search term in the centre and related concepts are shown in a circular visualization categorised by broad topics like "Persons", "Organizations", "Work" or "Society". Semantic relatedness between query term and related concepts are shown by the distance to the center; textual relations can be shown in a pop-up window.

---

2   Available at SeRE website: http://www.vizgr.org/sere/.
3   EyePlorer is a graphical knowledge engine. It provides an easy to use interface for exploring and interacting with a database of structured knowledge that contains more than 160 million facts. Available at: http://*www.vionto.com*.

Different visualization techniques have also been used for the presentation of search engine results. Treharne & Powers (2009) provide a useful literature summary and categorisation of retrieval systems and visual search engines such as *EyePlorer, Kartoo or Grokker* and offer an overview of reduced spatialisation techniques for search result visualization. Börner & Chen (2002) give an overview of usage scenarios of visualization for Digital Libraries like visual interfaces for searching and browsing, to get an overview of the entire document collection and for the visualization of user interaction data. In a similar way, Boulos (2003) presents an outline of visual maps for presenting and browsing purposes of large document collections in the domain of health information. Finally, Koshman (2006) gives an overview of visualization techniques and visual systems such as *Kartoo, Grokker or MapStan* for visualization-based information retrieval on the Web based on theories of human perception.

Wikipedia is a rich resource for concepts, semantic relations, facts and descriptions, which has been used in several research areas like natural language processing, information retrieval, information extraction and ontology building (Medelyan et al., 2009). *YAGO* (Suchanek, Kasneci & Weikum, 2007) and *DBpedia* (Auer et al., 2007) make use of structured information available in Wikipedia's infoboxes and Wikipedia's category system to build large knowledge bases. These resources can be used as semantic data sources to extract and compute related concepts around a chosen topic.

Wikipedia and the semantic databases built on it (notably *DBpedia* and *YAGO*) have been frequently used for the visualization of relationships between concepts or the hierarchical access via facets. Such an example is *RelFinder* - an interactive application that shows semantic relationships between two or more concepts in a graph visualization (Heim, Ertl & Ziegler, 2010). Users can explore existing relationships and can get additional information on concepts and semantic relationships. Faceted browsers make use of orthogonal properties of entities to browse, filter or query semantic databases. For the visualization part, different types like graph visualization, maps or hierarchical text filtering can be used. For instance, *Oobian Insight*[4] let users drill down results for *DBpedia* concepts in a graph, textual or a map view. Similarly, *gFacet* uses a network graph to browse the Web of data (Heim, Ziegler & Lohmann, 2008) or to construct complex semantic queries visually (Heim, Lohmann & Stegemann, 2010).

---

4   Available at http://dbpedia.oobian.com.

## 3. Computing semantically related concepts

In this chapter we explain how we compute semantically related concepts for a given search term and enrich them with additional information, most typically, categorisation, text snippets and thumbnails. The computation is encapsulated in a web service that returns all gathered information in an XML format. The XML file is then used by the web application as a data base. The web service is controlled by URL parameters. A search term can be passed and output fields can be specified. The algorithm then works as follows:

As a first step, the algorithm determines the best fitting Wikipedia concept for the search term. This is done with the help of the Wikipedia query module, which returns a list of the ten most relevant articles for the search term. We take the first entry as a corresponding Wikipedia concept for the given search term.

Then, related terms for this concept like in-/out-links, broader/narrower terms and categories are queried from online sources like Wikipedia and *DBpedia*. The Wikipedia data source is queried via the Wikipedia API and returns in- and out-links. *DBpedia* is queried via the SPARQL endpoint and returns category information, broader and narrower terms for the given Wikipedia concept. All these terms are ordered in an array and are used for the next step.

For each of the terms we compute the semantic relatedness (SR) to the original Wikipedia article. For doing so, we use our own adapted measure based on the Normalized Google Distance (NGD) (Cilibrasi & Vitanyi, 2007). The original NGD uses search hit counts from Google to compute the semantic relatedness between two terms. In our case, we use hit counts from the Wikipedia full text search to compute SR. The adapted Wikipedia Normalized Distance (WND) is computed with the following formula:

$$SR = \frac{\log_{10}(\max(A,B)) - \log_{10}(A \cup B)}{\log_{10}(W) - \log_{10}(\min(A,B))}$$

$A$: Number of full text search hits in Wikipedia for concept one

$B$: Number of full text search hits in Wikipedia for concept two

$A \cup B$: Number of full text search hits in Wikipedia for concept one AND concept two.

$W$: Number of articles in Wikipedia.

We set the SR value to 0 if $A \cup B = 0$, which means that the search does not return any results for both terms. If a concept has a SR value above 0, we additionally query for its category information, a thumbnail from Wikipedia or DBpedia and text snippets from the Wikipedia full text search.

Since every concept has several categories in *DBpedia*, we have to determine the categories with the most entries for a consistent overall classification. First, we order each concept in their category groups and sort groups by their number of entries. Then, the group with the most entries for a concept is assigned to it. With only one category for each concept, we are later able to sort the results for a query term, first by the number of entries in a category and then, inside the group, by SR values.

Text snippets are used to describe the relation between the original query term and the concepts found. To compute these snippets, we use a two-track approach: First, the full text of the original Wikipedia concepts is searched for the link and, if found, the sentence is extracted. Second, if the link is not found in the full text (i.e. because it was originally a broader term or a term from categorisation and not included in the full text) we use the full text search API of Wikipedia to search for a Boolean AND-combination of original and related concept and then use the resulting text snippets.

As a result, for an arbitrary search term we have a ranked list with a high number of semantically related concepts with their SR values, the most common category, a thumbnail and text snippets describing the relation to the search term. All these processing steps are computed live, in a parallel manner, with several hundred queries in parallel. Results for a search term can be computed in less than a few seconds. To further speed up the process, a caching mechanism can hold already computed results. This way, the system is able to be included in an interactive user interface to let the user query different search terms in all languages and explore their relations.

**4. User Interface**

In this section we would like to share some insights we gained in the process of the user interface (UI) development. Obviously, the goal of the UI was to let the user explore related concepts and their connections to a search term in a simple, joyful and interactive way.

In the first version, we experimented with a circular layout similar to the one used in *EyePlorer*. The idea was to reuse the graphical metaphor of proximity to show semantic relatedness. However, we quickly realized that in a radial design screen space is not sufficient to display, for example, a high number of concepts, categories or additional information like thumbnails. Of course, we could limit the number of related concepts and categories, but this would require more user interaction for choosing categories, viewing more concepts, or getting additional information for a concept.

In the second design we decided for a compromise between a standard result list and a graphical design. Here, we could combine the benefits of display forms using lean, ubiquitous, scalable, consistent, simple and intrinsic design of a rank-ordered list (Treharne & Powers, 2009) and the graphical variant with the visual encoding of Wikipedia concepts and their semantic relatedness to a search term.

Figure 1 shows the structure of the user interface. In the upper part, users can click the help button, change the overall language between English and German and can enter a query in the search form. An autocomplete function shows existing Wikipedia entities to facilitate the selection process. After the selection of a Wikipedia entity, an infobox shows a thumbnail, a link and a description of the chosen concept.

In the lower part, related entities are shown in the result list, ordered by their semantic relatedness to the search concept. For the visualization of concepts we use small panels with a thumbnail, a link to the Wikipedia article and a coloured marker at the top to visualize the semantic relatedness with colours from red to blue (more red = higher SR value to more blue = lower SR value). This way, semantic relatedness is encoded twice: (1) by the order of the panels and (2) by the colour marking. Hovering with the mouse pointer over a panel shows a popup window with text snippets describing the relation between the searched concept and the actual concept and with links to the related Wikipedia articles. With the help of a select box users can filter results. Here, the category information is used to let the user filter related concepts by categories that are available for the specific search term.

The application is multilingual and can be used in all different languages in which a Wikipedia version is available (in the current version we use English and German). Since Wikipedia articles in different languages are built up differently, with different text, links, categorisation etc., the same query concept can lead to different concepts, relations and categories to explore for different languages.

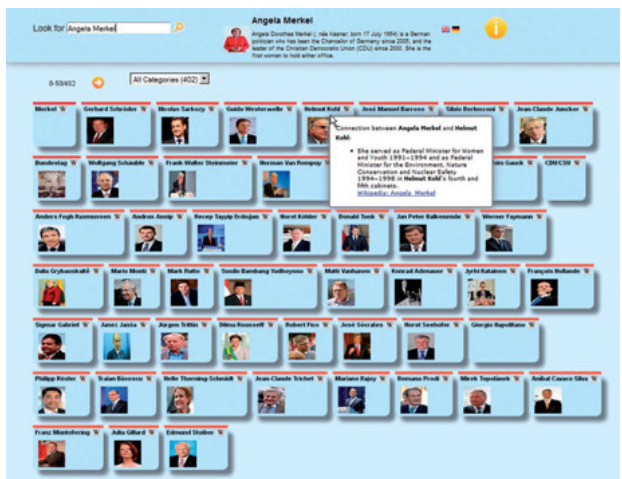

**Figure 1:** *A screenshot of SeRE with a search for the German Chancellor Angela Merkel and her connection to Helmut Kohl*

## 5. Evaluation

In a user test we evaluated how the users interact with the interface of the SeRE web application and examined the following research questions:

- Can the participants use the interface of SeRE intuitively?
- How do the users search with Google for related concepts?
- What are the differences between searching important entities and their relationships for a given topic with SeRE or Google?
- Is the sorting of related topics by semantic relatedness in SeRE helpful?
- How do the respondents evaluate and compare the two search strategies and which do they prefer?

### 5.1 Method & participants

The participants were asked to carry out a set of tasks in Google and in the prototype, and to fill out a questionnaire after each task in order to capture the actions performed, results found, observed difficulty level and comments.

Participants should first provide some personal information (gender, age, education degree and years of employment) and a self-assessment on a five-point-scale about their experience in information search on the Web. In the first part of the questionnaire, the users were asked to complete three different tasks based on the Google search. After that, they were prompted to familiarize themselves with the environment of SeRE for five minutes. In the second part, the participants had to resolve the same tasks as in part one, but now with the SeRE tool. Finally, the users were asked to compare and to evaluate the positive and negative aspects of the different search strategies, to state which they would prefer, and finally to assess the overall scenario.

The group of participants (n=9) included 8 male users and one female between the age of 26 to 40 (mean: 30 years). 8 participants were researchers and had a graduate degree in computer science such as M.Sc. or similar and one was an IT specialist. The rating about their own experience in dealing with web-based search was 1.44 ("very good").

### 5.2 Tasks & Questions

The participants had to handle the following tasks and afterwards they were asked to fill in the questionnaire. The first three tasks were the same for Google and SeRE search, starting to process them first with the Google search, then with SeRE. Each task should be processed within five minutes. For the Google search, users should additionally specify the source of information they used.

*1. Relevance of persons*

Participants should search and find five persons who played a major role in the political career of Angela Merkel. They were asked to write them down in descending order, to add a confidence score on a five-point-scale (1=very unsure, 2=unsure, 3=normal, 4=sure, 5=very sure) and to assess how difficult the task was perceived on another five-point-scale (2=very easy, 1=easy, 0=normal, -1=difficult, -2=very difficult). Finally, they could give comments, suggestions and criticism.

*2. Relations between persons*

The second task was to find information about possible relations of Angela Merkel and Jean-Claude Juncker. Similar to task 1, they had to record a confidence score for each answer on a five-point-scale, to assess the difficulty and were able to give comments and suggestions.

*3. Role of banks in the euro crisis*

In the third task, users should cite five most important banks in the context of the current euro crisis. Additional information like confidence score, difficulty, comments and suggestions are analogous to task 1 and 2.

*4. Comparison and evaluation of the two search strategies in the three tasks and overall results*

The overall scenario is then evaluated:

- What are the positive and negative sides of both search methods for the three tasks?
- Which search strategy would you favour for the given three tasks and why would you do so?
- How do you assess on a five-point-scale (2=very helpful, 1=helpful, 0=normal, -1=not helpful, -2=not helpful at all) the interacting technique, the visualization of data and your satisfaction with the results in the SeRE tool?
- Was the ranking of search results by semantic relatedness helpful for finding answers to the three tasks? Why and why not?

Finally, SeRE could be rated on a five-point-scale (2=very helpful, 1=helpful, 0=normal, -1=not helpful, -2=not helpful at all) and there was a space provided for general comments and suggestions.

### 5.3 Results

*1. Relevance of persons*

For Google, search participants named persons like Helmut Kohl, Wolfgang Schäuble or Lothar de Maizière (see Table 1 for comparison) who have played an important role in the political career of Angela Merkel. The noted sources were mainly Wikipedia, followed by some newspapers and a few other websites. The users recorded a confidence score on average with "sure" and the difficulty level with "normal". Two participants criticized the short time of 5 minutes to process the task; one mentioned that it was difficult to avoid search results from Wikipedia. For SeRE, the participants could use the interface intuitively and named persons like Christian Wulff, Helmut Kohl or Franz Müntefering. The confidence score was lower with "normal" and the difficulty still "normal". One user noticed that some explanations in the snippets were incomplete.

*2. Relations between persons*

In task 2 of the Google search participants reported relations of Angela Merkel and Jean-Claude Juncker such as "several topics referring to the euro crisis", "election support from Juncker" or "same party affiliation". In contrast to task 1 newspaper websites dominated as information sources, while Wikipedia was mentioned only twice. On an average, the participants were sure about the information they have found on the Web. The difficulty was evaluated as "normal". With the SeRE application, the majority named two or three relations between Merkel and Juncker such as "owner of Karlspreis" or "member of Frankfurter Runde". They had an average confidence in their search results with a difficulty perceived as normal. One subject criticized that a search for two names is not possible in *SeRE* and another remarked that the snippets showing the relationships were too short.

*3. Role of banks in the euro crisis*

The participants seemed to have more problems in solving task 3 with Google search. A majority cited 4 or 5 important banks like EZB, Lehmann Brothers or Commerzbank. They were tendentially confident in their answers (3.89), but assessed the difficulty of this task on average with "difficult" (-0.67). The sources of their results ranged evenly distributed from Wikipedia to different newspaper and television websites. With the SeRE application, participants stated banks like EZB, Deutsche Bundesbank or Lehmann Brothers. The confidence value of their search results were on average "normal" and assessed the difficulty level with "normal" and a slight tendency to "difficult" (-0.44). One subject wished that the concepts found could be directly used as search terms. Participants stated that the auto-complete function was very helpful and the most important banks were presented directly at a glance.

**Table 1:** *Found answers for Task 1 to 3, A= absolute answers, C=confidence scores (1=very unsure to 5=very sure)*

| Task | Google | A | C | SeRE | A | C |
|---|---|---|---|---|---|---|
| 1: Five important persons that played a major role in the political career of Merkel | 1. Helmut Kohl | 7 | 4.57 | Christian Wulff | 6 | 3.16 |
| | 2. Wolfgang Schäuble | 7 | 4.28 | Helmut Kohl (1.) | 3 | 3.33 |
| | 3. Lothar de Maizière | 5 | 3.4 | Franz Müntefering | 3 | 3.33 |
| | 4. Gerhard Schröder | 2 | 4 | Nicolas Sarkozy | 2 | 3.5 |
| | 5. Edmund Stoiber | 2 | 2 | Gerhard Schröder (4.) | 2 | 2.5 |
| 2: Relations between Merkel and Juncker | Topics referring to euro crisis | 5 | 4.2 | Karlspreis | 6 | 2.5 |
| | Juncker supported Merkel, e.g. in elections | 6 | 4.6 | Frankfurter Runde | 5 | 4 |
| | Party affiliation | 1 | 4 | Christine Lagarde | 1 | 4 |
| | | | | Hermann van Rompuy | 1 | 4 |
| | | | | José Manuel Barroso | 1 | 4 |
| 3: Five important banks in the euro crisis | 1 EZB | 5 | 4.2 | EZB (1.) | 8 | 3.9 |
| | 2. Lehmann Brothers | 3 | 4.6 | Deutsche Bundesbank (4.) | 5 | 3 |
| | 3. Commerzbank | 3 | 4.3 | Lehmann Brothers (2.) | 3 | 5 |
| | 4. Deutsche Bank | 3 | 4 | Banco de Portugal | 4 | 2 |
| | 5. Goldmann Sachs | 2 | 4 | Bank of England | 3 | 2.6 |

## 4. Comparison and evaluation of the two search strategies in the three tasks and the overall results

For the Google search the subjects perceived as positive the fact that there were broader data sources and different sources available and that they could use the search terms in different combinations. Furthermore, they described positively that text information was presented at a glance, snippets could be seen directly and that they obtained more extensive information. In contrast, they qualified as negative that there were no concrete concepts but only websites, that there was a lot of redundancy and that results could not be filtered according to special categories. Additionally, one user criticized that it was difficult to search for related entities.

The users noticed as advantage for the search with *SeRE* that there is no redundancy discernible, that there exists a good presentation of the results at a glance and that they obtained results sorted by semantic relatedness. Furthermore, they found the snippets helpful and found that it was easy to search for related entities. In most cases, the snippets showed the relationship between entities. Nearly all subjects criticized that *SeRE* uses only Wikipedia as a search basis. A few pointed out that the snippets were partially too short, could be seen only during mouse-over action and that search terms could not be combined. One person asked if the sorting of search results by semantic relatedness is always correct. Finally, some commented that the relevance on the snippets was almost always marked red, meaning a high semantic relatedness to the search term.

Three users preferred definitely Google search for the three tasks, because this search strategy was more helpful and powerful for solving the use cases. The other participants made their decision in a different way: They would like to combine both search strategies depending on the task. One subject, for example, would apply *SeRE* for the tasks 1 and 2, because relationships were directly visible, and Google for task 3 as it provides necessary background information to the issues. Another user would favor Google search basically for connections between two entities, but *SeRE* for looking for important entities for a search term.

Three users approved basically the sorting of search results by semantic relatedness, but in some cases the colour-coding of semantic relatedness was not fine-grained enough and there were too many irrelevant hits. Other three subjects found the sorting only partially helpful, because the relevance ranking between the results was not always evident. Two participants considered without detailed explanation the sorting by semantic relatedness as not particularly helpful for solving the three tasks. Table 2 shows detailed results of the evaluation.

**Table 2:** *Detailed results*

| Task | Google (average, standard deviation) | SeRE (absolute, standard deviation) |
|---|---|---|
| 1: Important persons – Merkel (absolute, average) | (40, 4.44) | (39, 4.33) |
| Confidence | "sure" (4.05, 0.93) | "normal" (3.18, 1.18) |
| Difficulty | "normal" (0.44, 0.73) | "normal" (-0.44, 1.24) |
| 2: Relations between Merkel – Juncker (absolute, average) | (25, 2.77) | (18, 2) |
| Confidence | "sure" (4.20, 0.96) | "normal" (3.44, 1.15) |
| Difficulty | "normal" (0.33, 0.87) | "normal" (0.00, 1.00) |
| 3: Important banks in the euro crisis (absolute, average) | (37, 4.11) | (35, 3.88) |
| Confidence | "normal" (3.89, 0.94) | "normal" (3.46, 1.40) |
| Difficulty | "normal" (-0.67, 0.87) | "normal" (-0.44, 1.13) |
| Final evaluation | | "normal" (0.33, 1.00) |
| Sorting of search results by semantic relatedness | | "normal" (-0.22, 0.97) |

## 6. Discussion & Conclusion

The comparative evaluation yielded interesting results for the task of entity and relationship visualization from Wikipedia. Google got positive scores for its large and rich data collection, arbitrary search term combinations, meaningful snippets including relationships, fast access and for the fact that users were familiar with its special interaction process. With SeRE we tried a new search approach with the reduction to ranked concepts and the database of Wikipedia. Results for task 1 (related persons) and 3 (important banks) are more or less comparable. Main issues seemed to be the ranking of concepts in SeRE and the provided relationships. Users asked about the reasons behind certain ranking of related concepts. This issue is directly connected to the presentation of relationships, which is explanatory for the underlying reasons. The main deficit of SeRE seems to be the diversity and quality of shown relationships. These can be deduced from the comments, but also from the confidence scores, which have been much higher for Google than for SeRE. We only used the Wikipedia database and tried to find phrases which contain both entities similar to the Google text snippets approach. However, phrases could not be found for all combinations. Alternatively, snippets showed Wikipedia articles which contain both entities, but often with much text in-between. The relationships were not directly visible in the snippets, but users must look into the text of the article and read details there. In a next version of SeRE, these relations should be obtained from a broader data corpus and directly integrated. This would combine the concept-oriented approach of SeRE with the power of natural language relationships from a broad database like Google.